%% file: main.tex
\documentclass[10pt,twocolumn,letterpaper]{article}

\usepackage[pagenumbers]{cvpr} 
\usepackage{amsmath,amssymb}
\usepackage{epsfig}
\usepackage{graphicx}
\usepackage{algorithm}
\usepackage{algorithmic}
\usepackage{setspace}
\input{preamble}

\usepackage[misc]{ifsym}
\definecolor{cvprblue}{rgb}{0.21,0.49,0.74}
\usepackage[pagebackref,breaklinks,colorlinks,citecolor=cvprblue]{hyperref}

\def\paperID{11193} 
\def\confName{CVPR}
\def\confYear{2024}

\def\METHODNAME{\textsc{MVOC}}

\title{\METHODNAME: a training-free multiple video object composition method \\ with diffusion models\vspace{-2mm}}
\author{Wei Wang$^{1*}$ \hspace{12pt} Yaosen Chen$^{1,2*{~\textrm{\Letter} }}$ \hspace{12pt}  Yuegen Liu$^2$   \hspace{12pt} Qi Yuan$^{1}$  \hspace{12pt} Shubin Yang$^{1,2}$ \hspace{12pt}Yanru Zhang$^{2}$ \\
	\normalsize{$^1 $Sobey Media Intelligence Laboratory \hspace{12pt}
		$^2 $University of Electronic Science and Technology of China}\\
	{\tt\small \{wangwei,chenyaosen,liuyuegen,yuanqi,yangshubin\}@sobey.com}\hspace{12pt}{\tt\small yanruzhang@uestc.edu.cn}\\
}

\begin{document}
	
	\twocolumn[{%
		\renewcommand\twocolumn[1][]{#1}%
		\maketitle
		\vspace{-2em}
		\begin{center}
			\centering

			\includegraphics[width=0.95\textwidth]{{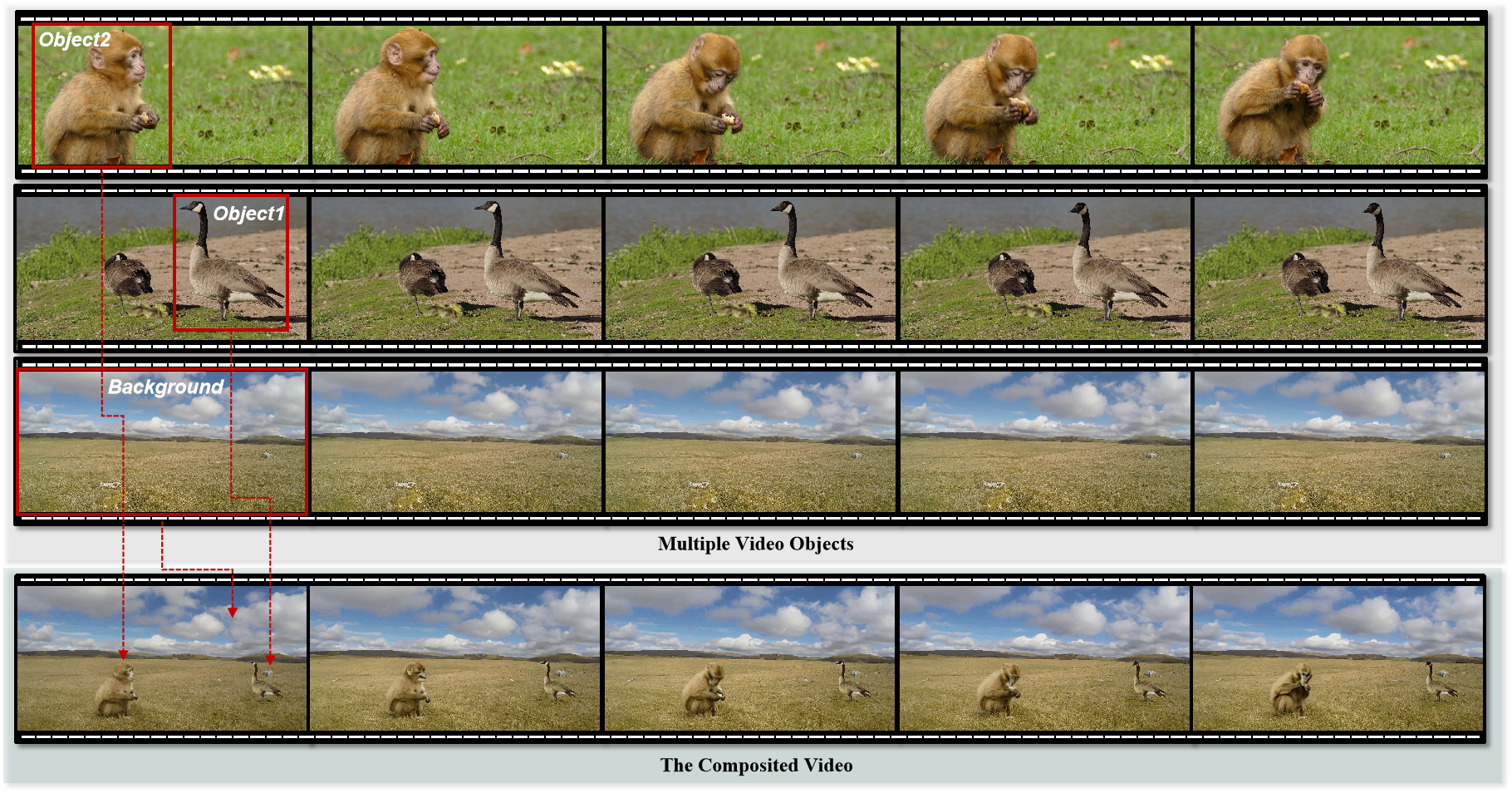}}
			\captionof{figure}{\textbf{Multiple Video Object Composition Results.} Given multiple video objects (e.g. Background, Object1, Object2),  our method enables presenting the \textbf{interaction effects} between multiple video objects and maintaining the \textbf{motion} and \textbf{identity consistency} of each object in the composited video.}
			\label{fig:results_overview}

		\end{center}%
	}]
	
\footnotetext{$*$ Equal Contribution. $~\textrm{\Letter}$ Corresponding Author.}

\input{sec/0_abstract}    

\input{sec/mvoc_introduction}
\input{sec/mvoc_related_work}
\input{sec/mvoc_preliminary}

\input{sec/mvoc_methods}

\input{sec/mvoc_experiments}

	
{
	\small
	\bibliographystyle{ieeenat_fullname}
	\bibliography{main}
}
	

\input{sec/mvoc_derivation}

\end{document}

%% file: preamble.tex
\usepackage[dvipsnames]{xcolor}



%% file: sec/0_abstract.tex
\begin{abstract}
Video composition is the core task of video editing. Although image composition based on diffusion models has been highly successful, it is not straightforward to extend the achievement to video object composition tasks, which not only exhibit corresponding interaction effects but also ensure that the objects in the composited video maintain motion and identity consistency, which is necessary to composite a physical harmony video. To address this challenge, we propose a Multiple Video Object Composition (MVOC) method based on diffusion models. Specifically, we first perform DDIM inversion on each video object to obtain the corresponding noise features. Secondly, we combine and edit each object by image editing methods to obtain the first frame of the composited video. Finally, we use the image-to-video generation model to composite the video with feature and attention injections in the Video Object Dependence Module, which is a training-free conditional guidance operation for video generation, and enables the coordination of features and attention maps between various objects that can be non-independent in the composited video. The final generative model not only constrains the objects in the generated video to be consistent with the original object motion and identity, but also introduces interaction effects between objects. Extensive experiments have demonstrated that the proposed method outperforms existing state-of-the-art approaches. Project page: \href{https://sobeymil.github.io/mvoc.com}{https://sobeymil.github.io/mvoc.com}.
\end{abstract}

%% file: sec/mvoc_introduction.tex
\section{Introduction}
\label{Introduction}

The role of media production and editing in the creation of engaging visuals for movies, short videos, and other forms of media is pervasive~\cite{chen2021boundary,chen2022video,chen2021capsule,chen2023hyperlips}. Multiple video object composition has long been a central tenet in the field of media production and editing. In this task, the objective is to composite new videos with objects from multiple input source videos in a manner that aligns with user intentions, visual aesthetics, and physical laws. In order to achieve this goal, the usual approach is to firstly separate the main objects of multiple source videos, then perform operations such as cropping, color retouching~\cite{chen2023nlut}, and mapping on each video object, finally optimize the edge details between video objects through compositing or blending~\cite{lu2021bridging,zhang2021deep,wu2019gp}. However, it is difficult to obtain a harmonious composited video through these methods.

Recently, conditional diffusion models have made a significant impact on image and video generation tasks~\cite{ramesh2022hierarchical,yuan2024magictime,bao2024vidu,blattmann2023stable,zhang2023i2vgen}, revolutionizing the field of visual creation. Diffusion models are usually guided by textual descriptions or images as conditions to generate images or videos that match the conditional semantics. A natural idea is whether it is possible to use diffusion models for object composition. Li, et al.~\cite{li2024tuning} proposed an approach that preserves the semantic features of the reference image subject while allowing modification of detailed attributes based on text descriptions. InteractiveVideo~\cite{zhang2024interactivevideo} introduces a synergistic multimodal instruction mechanism to seamlessly integrate generative models with user multimodal instructions, which can be regarded as objects.  Furthermore, IMPRINT~\cite{song2024imprint}, TF-ICON~\cite{lu2023tf}, PrimeComposer~\cite{wang2024primecomposer}, etc. all utilize the diffusion generation model to achieve the synthesis of image objects, and achieve good results, but most of them composite text or image objects, not video objects. 

To address the video composition issue, Guo et al.~\cite{guo2024training} proposed inter-frame augmented attention based on the image-based diffusion generation model to composite one object video and a background video. Although this method can maintain color consistency between the object and background videos, it is unable to demonstrate the interaction effect between them. This is largely due to the fact that the fundamental model employed is based on the diffusion model for image generation, which exhibits limitations in temporal consistency and multiple video object interactive effects. In contrast to this method, the pre-trained video-based generative diffusion model used in our method is inherently object motion-consistent. Liu et al.~\cite{liu2022compositional} have proposed a method for compositional visual generation with composable diffusion models. Inspired by them, we introduce a novel Multiple Video Object Composition pipeline. Furthermore, AnyV2V\cite{ku2024anyv2v} suggests a training-free video editing approach, which motivates us to introduce the Video Object Dependence module, which considers feature and attention injections. Consequently, effects generated by object interactions can be well handled, and the composition of multiple non-independent video objets can be supported. Fig.~\ref{fig:results_overview} illustrates the multiple video object composition results with our method. Our contributions are summarized as follows:
\begin{itemize}[leftmargin=0.1cm, itemindent=0.1cm]
	\item We propose a training-free pipeline for composing multiple non-independent video objects to handle object interactions  and generate harmonious videos.
	\item We propose the Video Object Dependence module, which involves video objects as conditions to maintain object identity and motion consistency.
	\item Extensive experiments have demonstrated that our method is capable of not only maintaining the \textbf{motion} and \textbf{identity consistency} of each object but also presenting the \textbf{interaction effects} between multiple video objects in the composited video.
\end{itemize}

%% file: sec/mvoc_related_work.tex
\section{Related work}
\label{related_work}

\noindent\textbf{Video generation models.}
The progress of diffusion models~\cite{sohl2015deep,ho2020denoising,rombach2022high} has promoted the great development in the field of visual generation, and various solution forms have evolved, including pixel-space diffusion models~\cite{ge2023preserve,ho2022imagen,ho2022video}, latent diffusion models (LDMs)~\cite{rombach2022high,zhou2022magicvideo,an2023latent}, and transformer diffusion models~\cite{lu2024fit,ma2024latte,peebles2023scalable}. The LDM-based video generation models have become the most widely studied methods due to its high efficiency and open-source stable diffusion~\cite{rombach2022high,blattmann2023stable} models. Introducing temporal modules into image generation models to ensure temporal consistency has become the main paradigm in video generation model research. The main advantage of these models is that it can use excellent open-source image generation models that have been widely proven as a basis to accelerate the research of video generation models~\cite{blattmann2023align,liew2023magicedit,shi2023bivdiff,an2023latent,zhang2023i2vgen}. Text-to-Video (T2V)~\cite{yuan2024magictime,bao2024vidu} and Image-to-Video (I2V)~\cite{blattmann2023stable,zhang2023i2vgen} models have become the fundamental models for generative-based Video-to-Video (V2V) editing methods. Consequently, the development of controllable video generation based on these fundamental models has emerged as a significant area of research.

\noindent\textbf{Generative video editing.}
The introduction of a series of additional conditions to achieve precise editing of original videos has broad application prospects.  For example, VideoComposer~\cite{wang2024videocomposer} decomposes video generation into texture conditions, spatial conditions, and temporal conditions, thereby achieving controllable motion video synthesis. CTRL-Adapter~\cite{lin2024ctrl} trains an Adapter which can support a variety of useful applications, including video control with multiple conditions, image control, zero-shot transfer and video editing. FRESCO~\cite{yang2024fresco} achieves zero-shot video translation by optimizing the UNet features. CustomVideo~\cite{wang2024customvideo} can generate identity-preserving videos with the guidance of multiple subjects. MOTIA~\cite{wang2024your} leverages both the intrinsic data-specific patterns of the source video and the image/video generation prior for effective outpainting. However, these methods all require time for training or fine-tuning to effectively implement video editing in actual production. On the contrary, some studies are investigating training-free methods to address this issue. ConditionVideo~\cite{peng2023conditionvideo} employs a pre-trained 3D control network to strengthen conditional generation accuracy by additionally leveraging the bi-directional frames in the temporal domain. BIVDiff~\cite{shi2023bivdiff} employs a framework that integrates frame-wise video generation, mixed inversion, and video temporal smoothing components for a training-free video synthesis. AnyV2V~\cite{ku2024anyv2v} utilizes an existing I2V generation model for DDIM~\cite{song2020denoising} inversion, as well as feature injections for V2V editing.


%% file: sec/mvoc_preliminary.tex
\section{Preliminary}

\subsection{Diffusion models}

Diffusion models~\cite{sohl2015deep,ho2020denoising,rombach2022high} are a class of probabilistic generative models in which generation is modeled as an iterative denoising procedure. To recover the actual visual content, diffusion models progressively remove noise from an initial Gaussian noise. These models are founded on two random processes. The forward process is fixed to Markov chain that progressively adds Gaussian noise to the data. We can sample $x_t$ at an arbitrary time step $t$ in closed form:
\begin{equation}\label{eq:qxtx01}
	\begin{aligned}
		q( \boldsymbol{x}_t|\boldsymbol{x}_0)=  \mathcal{N}\bigl(\boldsymbol{x}_{t}; \sqrt{\overline{\alpha}_t}\boldsymbol{x}_0, (1- \overline{\alpha}_t)I\bigl),
	\end{aligned}
\end{equation}
where $\overline{\alpha}_t:=\prod_{s=1}^t\alpha_s$, is derived from the variance noise schedule. The reverse process begins from the noise and transitions towards the original data $q(\boldsymbol{x}_0)$. 

Denoising Diffusion Probabilistic Models (DDPM)\cite{ho2020denoising} are a subset of diffusion models. They choose a sequence of noise coefficients for Markov transition kernels following specific patterns, which are generally constant, linear and cosine schedule. DDPM ascertain a denoising function $\epsilon_\theta(\boldsymbol{x}_t, t)$ to estimate the noise added at each step as shown in Eq.~\ref{eq:qxtx0}. 

\begin{equation} \label{eq:qxtx0}
	\begin{aligned}
		p_{\theta}(\boldsymbol{x}_{t-1}|\boldsymbol{x}_{t})= \mathcal{N}\bigl(\boldsymbol{x}_{t};\frac{1}{\sqrt{\alpha_{t}}}(\boldsymbol{x}_{t}-\frac{1-\alpha_{t}}{\sqrt{1-\bar{\alpha}}_{t}}\epsilon_{\theta}(\boldsymbol{x}_{t},t)),\sigma^{2}_{t})	 
	\end{aligned}
\end{equation}

\subsection{DDIM inversion}

The denoising process for diffusion models from $x_t$ to $x_{t-1}$ can be achieved using the Denoising Diffusion Implicit Models (DDIM)~\cite{song2020denoising} sampling algorithm. Different from DDPM, DDIM directly derive the following relationship without relying on Markov chain.
\begin{equation} \label{eq:ddim_r111yce}
\begin{split}
		\boldsymbol{x}_{t-1}&=\sqrt{\bar{\alpha}_{t-1}}{\frac{\boldsymbol{x}_{t}-\sqrt{1-\bar{\alpha}_{t}}\epsilon_{\theta}(\boldsymbol{x}_{t},t)}{\sqrt{\bar{\alpha}_{t}}}}\\
		&+   {(\sqrt{1-\bar{\alpha}_{t-1}-\sigma_{t}^2})\epsilon_{\theta}(\boldsymbol{x}_{t},t)}+ {\sigma_{t}\varepsilon_1}, 
\end{split}
\end{equation}
where $\sigma_{t}$ is the variance and $\varepsilon_1$ is the added gaussian noise. DDIM has two major advantages: 1. Since Markov chain is not introduced in the derivation process, the sampling process represented by Eq.~\ref{eq:ddim_r111yce} is not strictly from $\boldsymbol{x}_{t}$ to $\boldsymbol{x}_{t-1}$, thereby realizing skip sampling and accelerating the corresponding generative processes; 2. If $\sigma_{t}$ is assigned $0$, the sampling process is uniquely determined and no noise is introduced. Consequently, we get the necessary condition for inversion.

The reverse process of DDIM sampling, known as DDIM inversion, allows obtaining $\boldsymbol{x}_{t}$ from $\boldsymbol{x}_{t-1}$ as follows.
\begin{equation} \label{eq:ddim_inversf}
	\begin{aligned}
		\boldsymbol{x}_t&= \sqrt{\frac{{\bar{\alpha}}_{t}}{{\bar{\alpha}_{t-1}}}}\boldsymbol{x}_{t-1}\\
		&+\sqrt{\bar{\alpha}_{t}}(\sqrt{\frac{1}{\bar{\alpha}_{t}}-1}-\sqrt{\frac{1}{\bar{\alpha}_{t-1}}-1})\epsilon_{\theta}(\boldsymbol{x}_{t},t)
	\end{aligned}
\end{equation}
In actual operation, $\epsilon_{\theta}(\boldsymbol{x}_{t},t)$ is approximated as $\epsilon_{\theta}(\boldsymbol{x}_{t-1},t-1)$, the inverted image $\boldsymbol{x}_{0}$ is brought in, and the noise $\boldsymbol{x}_{t}$ is gradually predicted.

\subsection{Attention mechanisms}
The denoising model $\epsilon_\theta(\boldsymbol{x}_t, t)$ is often instantiated by architectures like UNet~\cite{ronneberger2015u}, which has four basic modules: convolutional modules, spatial self-attention modules, spatial cross-attention modules, and temporal self-attention modules. In this work, we focus on latent diffusion models~\cite{Rombach_2022_CVPR} for video generation, which formulates the attention operation as:
\begin{subequations} \label{eq:prim_attention}
	\begin{align}
		Q=W^Qz,K=W^Kz,V=W^Vz, \label{eq:prim_attentiona}
		\\
		\text{Attention}(Q,K,V)=\text{Softmax}(\frac{QK^\top}{\sqrt{d}})V, \label{eq:prim_attentionb}
	\end{align}
\end{subequations}
where $z$ is the input hidden state to the attention layer and $W^Q$, $W^K$ and $W^V$ are trainable projection matrices that map $z$ onto query, key and value vectors, respectively. For spatial self-attention, $z$ represents a sequence of spatial tokens from each frame. For temporal self-attention, $z$ is composed of tokens located at the same spatial position for all frames of the video.

%% file: sec/mvoc_methods.tex
\begin{figure*}[t!]
	\centering
	\includegraphics[width=1.0\textwidth]{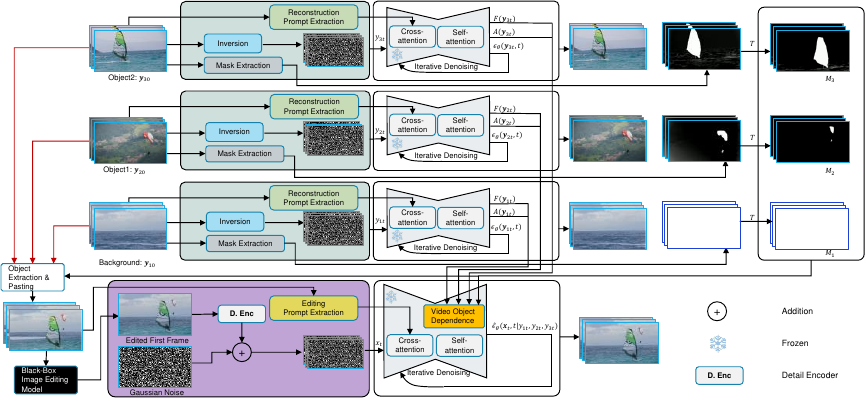}
	\caption{\textbf{Multiple video object composition framework.} Our method presents a two-stage approach: video object preprocessing and generative video editing. In the preprocessing stage, we perform DDIM inversion, object extraction and paste, as well as mask extraction. In the editing stage, we edit the first frame by an image editing model, then use video object dependence for conditional guidance video generation.}
	\label{fig:mvoc_framework}
\end{figure*}
\section{The proposed method}

\subsection{Overall framework}
The proposed  multiple video object composition framework is illustrated in Fig.~\ref{fig:mvoc_framework}. For convenience, we omit the description and illustration of autoencoder that compresses a RGB pixel space to a low-resolution latent space and reconstructs the latent space back to RGB frames in latent diffusion models. 

Our method presents a two-stage approach: video object preprocessing and generative video editing. In the first stage, given multiple video objects $\{\boldsymbol{y}_{i0}\}_{i=1}^N$, we first perform DDIM inversion to obtain the corresponding noise $\{{\boldsymbol{y}_{iT}}\}_{i=1}^N$. Meanwhile,  the video objects are derived to the corresponding object masks $\{\boldsymbol{M}_{\boldsymbol{y}_{i}}\}_{i=1}^N$ by the mask extraction module. 

In the second stage, we combine the video objects and the corresponding masks into a new conditional guidance video $\boldsymbol{x}_0$. Specifically, we leverage an image editing model to obtain an edited first frame $\hat{I}_1$, and feed it into the I2V generation model for video $\boldsymbol{x}_0$ generation. In this process, for depending on the original video objects, we use the features and attention maps of the corresponding noise $\{\{{\boldsymbol{y}_{it}}\}_{i=1}^N\}_{t=1}^T$ as condition injections to generate the composited video noise $\{\boldsymbol{x}_t\}_{t=1}^T$ (ref \ref{subsec:VideoObjectDependence}). The video objects may not be independent in the final video, which is generated by a novel multiple dependence generation method with diffusion models (ref \ref{subsec:multipleDependence}). For the $i$th video object, its features and attention maps can be denoted as $\{F({\boldsymbol{y}_{it}})\}_{t=1}^T$ and $\{A({\boldsymbol{y}_{it}})\}_{t=1}^T$, respectively.

\subsection{Multiple dependence generation through diffusion models}\label{subsec:multipleDependence}

We take a gradient of logarithm on both sides of Eq.\ref{eq:qxtx01} w.r.t $\boldsymbol{x}_t$:
\begin{equation}
    \begin{split}
       \nabla_{\boldsymbol{x}_t} \log q(\boldsymbol{x}_t |\boldsymbol{x}_0)=-\frac{\boldsymbol{x}_t-\sqrt{\overline{\alpha}_t}\boldsymbol{x}_0}{1-\overline{\alpha}_t}=-\frac{\epsilon}{\sqrt{1-\overline{\alpha}_t}}
     \end{split}
\end{equation}
$\epsilon$ is predicted by a neural network $\epsilon_\theta(\boldsymbol{x}_t,t)$, and then we have score function.

\begin{equation}\label{Eq:Socrefunction}
    \begin{split}
       \nabla_{\boldsymbol{x}_t} \log p(\boldsymbol{x}_t)=-\frac{\epsilon_\theta(\boldsymbol{x}_t,t)}{\sqrt{1-\overline{\alpha}_t}}
     \end{split}
\end{equation}

Supposing there are a set of video objects, which have noise $\{{\boldsymbol{y}_{it}}\}_{i=1}^N$ at each time step $t$, and the final composited video that have noise $\{\boldsymbol{x}_t\}_{t=1}^T$, we employ distinct denoising processes to recover them. At an arbitrary time step $t$, the probability distribution of $\boldsymbol{x}_t$ depends on $\{\boldsymbol{y}_{it}\}_{i=1}^N=\{\boldsymbol{y}_{1t}, \boldsymbol{y}_{2t}, \ldots, \boldsymbol{y}_{nt}\}$ as follows:
\begin{equation}\label{eq:Socrefunction_bayes}
    \begin{split}
        p(\boldsymbol{x}_t |\boldsymbol{y}_{1t}, \ldots, \boldsymbol{y}_{nt}) &\propto p(\boldsymbol{x}_t)\prod_{i=1}^n \frac{p(\boldsymbol{x}| \boldsymbol{y}_{1t}, \ldots, \boldsymbol{y}_{it})}{p(x | \boldsymbol{y}_{0t}, \ldots, \boldsymbol{y}_{(i-1)t})}
    \end{split}
\end{equation}
where $\boldsymbol{y}_{0t} = \phi$, which denotes the null condition. Then using the score function Eq.\ref{Eq:Socrefunction}, we get the composited noise prediction at time step $t$:
\begin{equation}\label{eq:condition_noise_p}
\begin{split}
    \hat{\epsilon}_{\theta}(\boldsymbol{x}_t, t |\boldsymbol{y}_{1t}, \ldots, \boldsymbol{y}_{nt}) &= 
    \epsilon_\theta(\boldsymbol{x}_t, t) \\
    &+ \sum_{i=1}^n w_i\bigl(\epsilon_\theta(\boldsymbol{x}_t, t | \boldsymbol{y}_{1t}, \ldots, \boldsymbol{y}_{it}) \\
    &- \epsilon_\theta(\boldsymbol{x}_t, t | \boldsymbol{y}_{0t}, \ldots, \boldsymbol{y}_{(i-1)t} )\bigl),
\end{split}
\end{equation}
where $w_i$ represents a set of hyper-parameters to control the dependency strength.

Given a set of independent objects $\{{\boldsymbol{y}_{it}}\}_{i=1}^N$, the above equation will reduce to the approach proposed by Liu et al.~\cite{liu2022compositional} as follows:
\begin{equation}\label{eq:independent condition_noise_p}
\begin{split}
    \hat{\epsilon}_{\theta}&(\boldsymbol{x}_t, t |\boldsymbol{y}_{1t}, \ldots, \boldsymbol{y}_{nt}) = 
    \epsilon_\theta(\boldsymbol{x}_t, t) \\
    &+ \sum_{i=1}^n w_i\bigl(\epsilon_\theta(\boldsymbol{x}_t, t | \boldsymbol{y}_{it})- \epsilon_\theta(\boldsymbol{x}_t, t)\bigl).
\end{split}
\end{equation}

In the setting where only one object is composed for sampling, the above equation will reduce to conditional generation, classifier-free guidance (CFG)\cite{ho2021classifier}:
\begin{equation}
    \hat{\epsilon}_\theta(\boldsymbol{x}_t, t | c) = \epsilon_\theta(\boldsymbol{x}_t, t) + w \bigl(\epsilon_\theta(\boldsymbol{x}_t, t| c) - \epsilon_{\theta}(\boldsymbol{x}_t, t)\bigl).
\end{equation}

\subsection{Training-free multiple video object dependence}\label{subsec:VideoObjectDependence}

For convenience, we can express Eq.\ref{eq:condition_noise_p} as follows by defining $\omega_i=w_i-w_{i+1}$,
\begin{equation}\label{eq:condition_noise_p_convience}
    \hat{\epsilon}_\theta(\boldsymbol{x}_t, t | \{\boldsymbol{y}_{it}\}_{i=1}^N) = \sum_{i=0}^n \omega_i\epsilon_\theta(\boldsymbol{x}_t, t| \{\boldsymbol{y}_{jt}\}_{j=0}^i),
\end{equation}
where $\epsilon_\theta(\boldsymbol{x}_t, t|\boldsymbol{y}_{0t})=\epsilon_\theta(\boldsymbol{x}_t, t)$, $w_0=1$, $w_{N+1}=0$.

\begin{figure}[t!]
	\centering
	\includegraphics[width=0.4\textwidth]{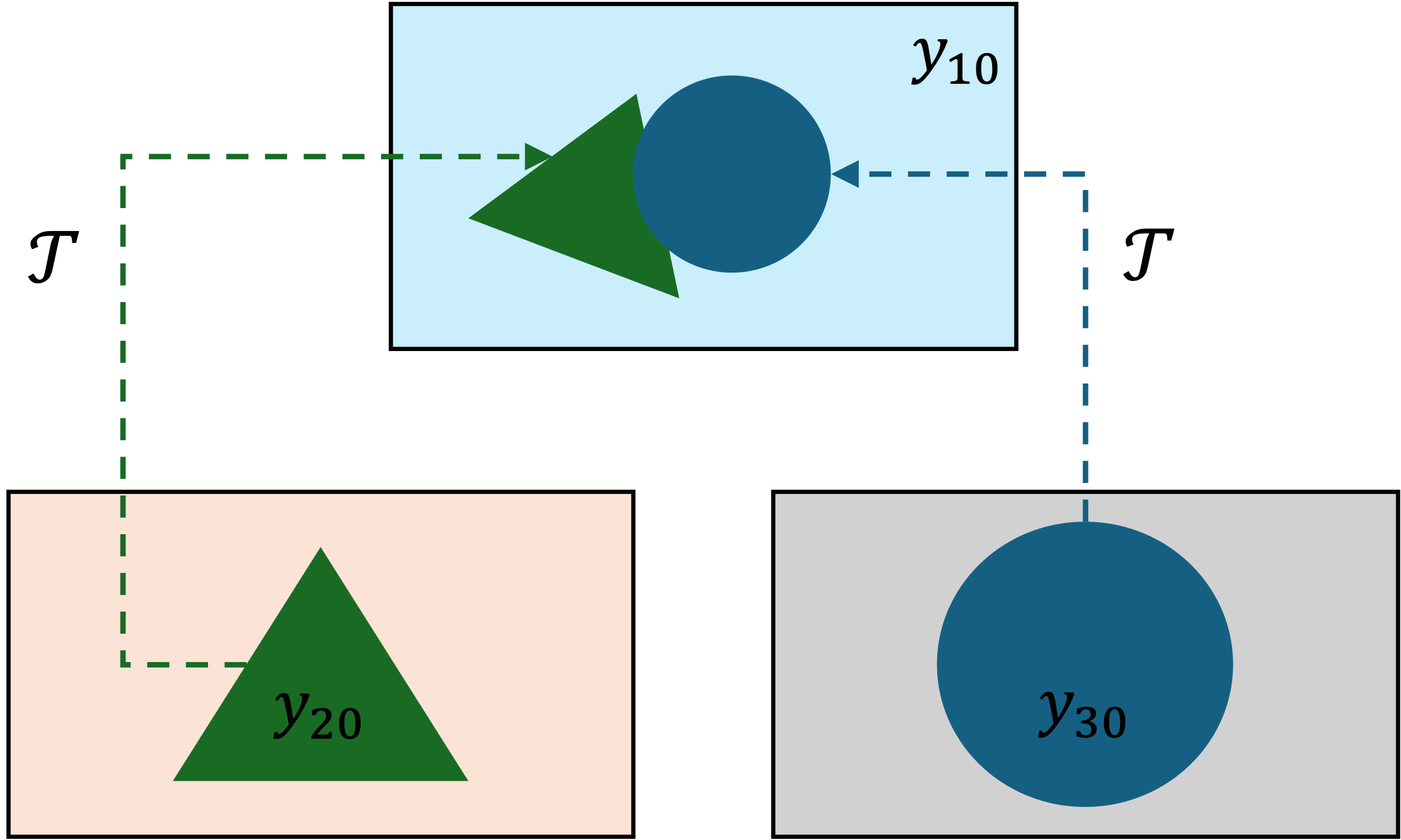}
	\caption{\textbf{Illustration of three video object composition.} } 
	\label{fig:vod}
\end{figure}

Aiming to get the right term in Eq.\ref{eq:condition_noise_p_convience}, we need control the video composition with multiple objects as constraints, which are generated by layered video objects one by one. For example, we consider placing two objects into one scene as shown in Fig.~\ref{fig:vod}. There are three objects in the final composited video, the background $\boldsymbol{y}_{10}$, two objects  $\boldsymbol{y}_{20}$ and  $\boldsymbol{y}_{30}$, which are layered combined together according to the distance of the camera.

For using denoising processes to generate composited videos with object dependences, we define the operations of feature extraction and attention map extraction in the models as $\mathcal{F}$ and $\mathcal{A}$, respectively. $\boldsymbol{F}_{0t}$ and $\boldsymbol{A}_{0t}$ are the features and attention maps of $\epsilon_{\theta}(\boldsymbol{x}_t, t)$, respectively. In other words, they are composited video noise features and attention maps without object dependence. From the camera view point, objects are layered on top of each other. Consequently, we get one object conditioned video has following features and attention maps.
\begin{subequations} 
	\begin{align}
		\begin{split}
		\boldsymbol{F}_{1t}&=\boldsymbol{F}_{0t}\odot(1-\boldsymbol{M}_{\boldsymbol{y}_{1}})+\mathcal{T}(\mathcal{F}(\boldsymbol{y}_{1t}))\odot\boldsymbol{M}_{\boldsymbol{y}_{1}},
	\end{split}
		\\
		\begin{split}
	\boldsymbol{A}_{1t}&=\boldsymbol{A}_{0t}\odot(1-\boldsymbol{M}_{\boldsymbol{y}_{1}})+\mathcal{T}(\mathcal{A}(\boldsymbol{y}_{1t}))\odot\boldsymbol{M}_{\boldsymbol{y}_{1}},
	\end{split}
	\end{align}
\end{subequations}
where $\odot$ denotes Hadamard Product, $\mathcal{T}$ denotes the affine transformation (ref Fig.~\ref{fig:vod}). Intuitively, we can calculate multiple objects conditioned features and attention maps layer by layer. Defining the features and attention maps by the first $i$ objects as $\boldsymbol{F}_{it}$ and $\boldsymbol{A}_{it}$ respectively, we can calculate them iteratively as Alg.\ref{algmultiple}.

\begin{algorithm}
\caption{Multiple object composition features and attention maps}
\begin{algorithmic}
\FOR {$i\in [1,N]$}
\STATE $\boldsymbol{F}_{it}=\boldsymbol{F}_{(i-1)t}\odot(1-\boldsymbol{M}_{\boldsymbol{y}_{i}})+\mathcal{T}(\mathcal{F}(\boldsymbol{y}_{it}))\odot\boldsymbol{M}_{\boldsymbol{y}_{i}}$
\STATE $\boldsymbol{A}_{it}=\boldsymbol{A}_{(i-1)t}\odot(1-\boldsymbol{M}_{\boldsymbol{y}_{i}})+\mathcal{T}(\mathcal{A}(\boldsymbol{y}_{it}))\odot\boldsymbol{M}_{\boldsymbol{y}_{i}}$
\ENDFOR
\end{algorithmic}
\label{algmultiple}
\end{algorithm}

Following \cite{ku2024anyv2v}, we use $\{\}$ represents feature and attention map injections, and then get 
\begin{equation}\label{eq:condition_injection}
    \epsilon_\theta(\boldsymbol{x}_t, t| \{\boldsymbol{y}_{jt}\}_{j=0}^i)= \epsilon_\theta(\boldsymbol{x}_t, t|\{\boldsymbol{F}_{it},\boldsymbol{A}_{it}\}).
\end{equation}

Finally, we can generate a video conditioned on multiple video objects by denoising processes with the following noise expression.
\begin{equation}\label{eq:multiply_condition_injection_final}
    \hat{\epsilon}_\theta(\boldsymbol{x}_t, t | \{\boldsymbol{y}_{it}\}_{i=1}^N) = \sum_{i=0}^n \omega_i \epsilon_\theta(\boldsymbol{x}_t, t|\{\boldsymbol{F}_{it},\boldsymbol{A}_{it}\})
\end{equation}

%% file: sec/mvoc_experiments.tex
\section{Experiments}\label{sec:experiments_4}
\subsection{Experimental setup}
\begin{figure*}[t!]
	\centering
	\includegraphics[width=1.0\textwidth]{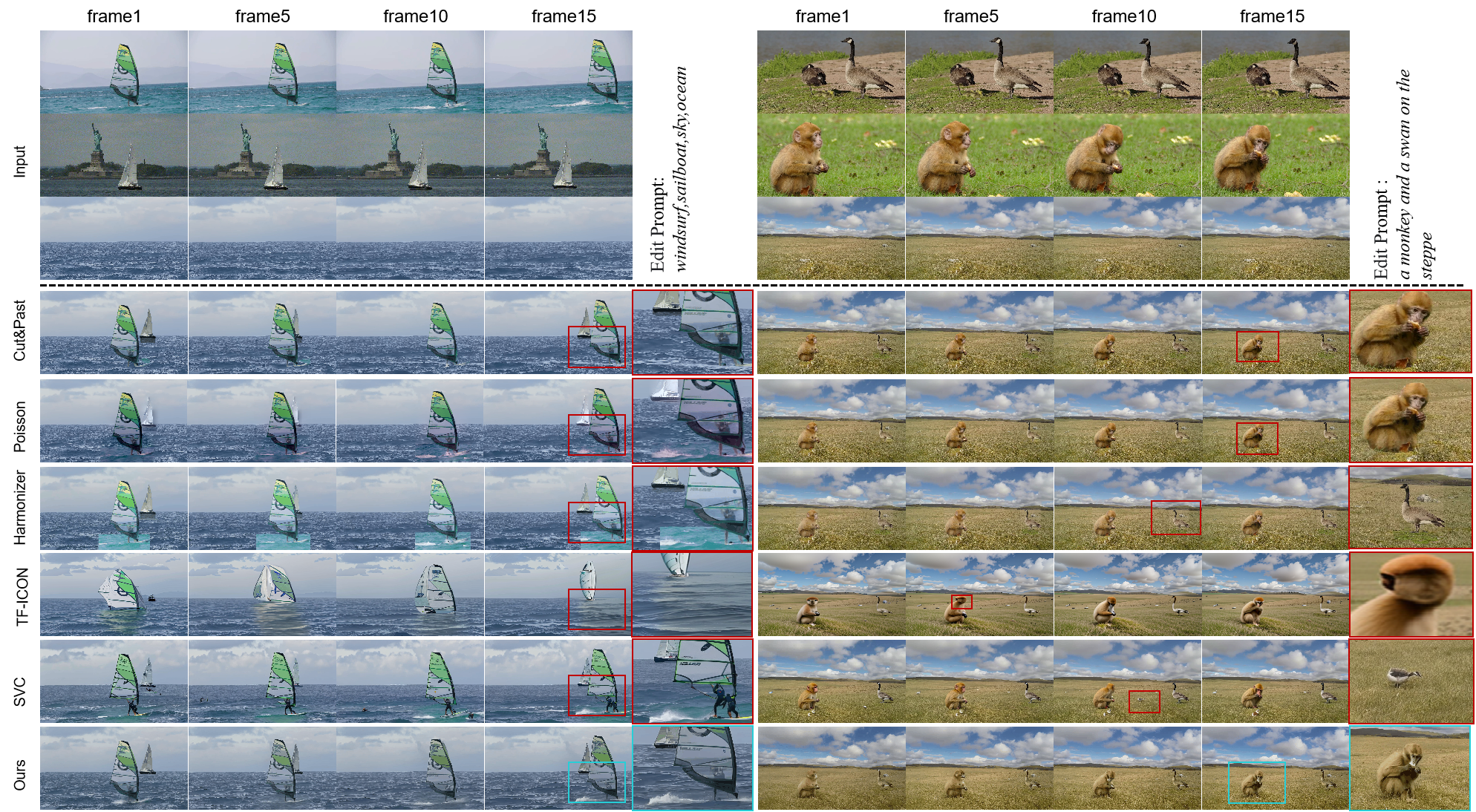}
	\caption{\textbf{Qualitative comparisons.} We utilize different methods to composite three video objects into one video. Our method is capable of not only maintaining the \textbf{motion} and \textbf{identity consistency} of each object but also presenting the \textbf{interaction effects} between multiple video objects in the composited video.
	}
	\label{fig:mvoc_visual_comparison}
\end{figure*}
We implement our Multiple Video Object Composition based on image-to-video (I2V)  I2VGenXL~\cite{zhang2023i2vgen} with image embedding~\cite{Radford2021LearningTV}. Seven groups of videos (two objects and one background) are collected from the Internet for evaluation and all videos are extracted to 16 frames and kept at a resolution of $1280\times720$. To obtain the edited first frame, we use the inpainting model in stable diffusion~\cite{rombach2022high}. We use Segment-and-Track-Anything~\cite{cheng2023segment} to extract the masks of moving video objects and manually select to extract the masks of objects that do not move significantly.

We composite the video in $N$ ($N=50$ in our experiments) sampling steps. There are two important operations in the Multiple Video Object Dependence module, i.e., feature and attention injections. The feature injection is divided into two parts, one part comes from the features before inputting the denoising UNet (denoted as $F_{n}$), and the other part comes from the output of all residual modules in the UNet (denoted as $F_{r}$).  The attention injection is also divided into two parts, one part comes from the temporal attention (denoted as  $A_{t}$), and the other part comes from the spatial attention (denoted as  $A_{s}$). We set the $r_{fn}$, $r_{fr}$, $r_{at}$, $r_{as}$ coefficients to denote the proportion of feature $F_{n}$, feature $F_{r}$, temporal attention $A_{t}$, and spatial attention $A_{s}$  injections in the full sampling process, respectively. For example, $r_{fn}=0.1$ means that the injection of $F_{n}$ is performed in only 5 steps, i.e., steps 1 to 5; $r_{fn}=0$ means that there is no feature injection of $F_{n}$ in all processes; in contrast, $r_{fn}=1$ indicates that feature injection was performed during all sampling. We empirically set $r_{fn}=0.02$, $r_{fr}=0.1$, $r_{at}=1$, $r_{as}=1$. 

All experiments are performed on a single NVIDIA A6000 GPU.



\subsection{Qualitative results}


 We compare our method with other state-of-the-art composition methods: 1) \textbf{CutPaste} composites the videos together directly through the mask of corresponding objects. 2) \textbf{Poisson}~\cite{perez2023poisson} utilizes generic interpolation machinery based on solving Poisson equations to achieve seamless area editing of images. 3) \textbf{Harmonizer}~\cite{ke2022harmonizer} treats image frame as an image-level regression task, learning the parameters of filters like humans do, and applying these filters to the original video to adjust the composite video. 4) \textbf{TF-ICON}~\cite{lu2023tf} is an image-based diffusion with cross-domain image-guided generative composition method. 5) \textbf{SVC}~\cite{guo2024training} is also an image-based diffusion method, which utilizes inter-frame augmentation attention to enhance the video object composition. Since SVC has not made their code public yet, we reproduced their method as described in the literature.
 
Fig.~\ref{fig:mvoc_visual_comparison} shows the visual comparison results. We take three video objects, i.e., background, object1 and object2, as input and composite them into one video with different methods. From the results, while Poisson and Harmonization, compared to CutPaste, show better color consistency among objects, they fail to create interaction effects between them. For example, the monkey and grass in the right clip are more rigidly blended. On the contrary, TF-ICON, SVC, and the proposed method are all based on generative methods, which can produce certain interactive effects, such as shadows. However, the objects in the TF-ICON composition video are not consistent with the objects in the original video. For example, the sailboat and monkey are distorted, and it is difficult to maintain the temporal consistency. SVC has better temporal consistency than TF-ICON, but the objects in the SVC composition video are still difficult to maintain the consistency with the original videos. For example, there is an extra person under the sail, and other objects are added to the grass where the monkey is located. It is clear that the proposed method performs better than the compared models in terms of temporal consistency, harmonization and object interaction effects. 
  
\begin{figure}[t!]
 	\centering
 	\includegraphics[width=0.45\textwidth]{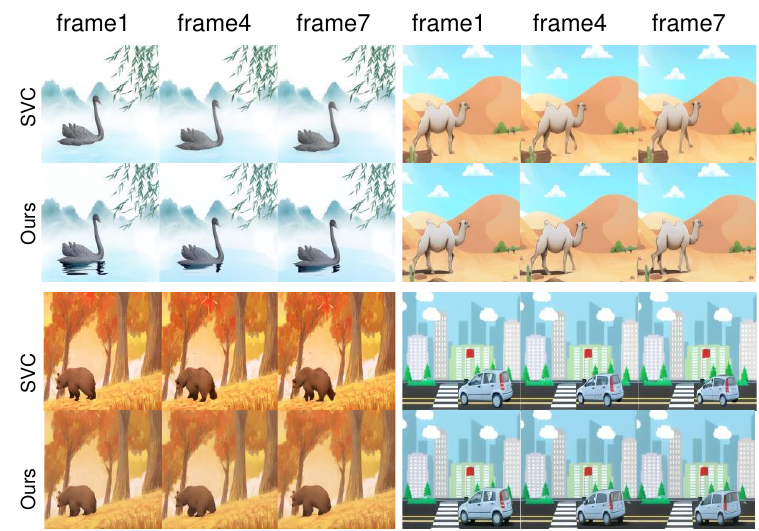}
 	\caption{\textbf{Qualitative comparisons with SVC.} Compared to SVC, our method performs better in generating interactive effects on objects, such as shadows. }
 	\label{fig:mvoc_visual_svc}
 \end{figure}

For a fair comparison, we also utilize the composited results publicly provided by SVC~\cite{guo2024training} in Fig.~\ref{fig:mvoc_visual_svc}, which shows the object composited by our method has a shadow in the new background under the cartoon scene while SVC can not produce them. It not only demonstrates the superior color consistency of our method but also presents the model ability to create interactive effects between objects and backgrounds.
 
\subsection{Quantitative results}
\noindent\textbf{Temporal consistency.} We use short and long-term consistency~\cite{zhang2018unreasonable,lai2018learning} to measure the temporal consistency of the composited videos. The consistency score formulates as: 
\begin{equation}
	E_{warp}(V_t, V_{t+G}) = \frac{1}{ \sum_{i=1}^{N} M^{(i)}_{t} }\sum_{i=1}^{N}M^{(i)}_{t}||V^{(i)}_{t}-\hat{V} ^{(i)}_{t+G}||^{2}_{2},
\end{equation}
where $\hat{V} ^{(i)}_{t+G}$ is the warped frame $V^{(i)}_{t}$ and $M_{t}^{(i)}$ is a non-occlusion mask estimated by the method in~\cite{ruder2016artistic}. $G$ is the number of interval frames.  
We compare the short and long-range consistency using the values computed at intervals of 2 frames ($V_{t},V_{t+2}$) and 4  frames ($V_{t},V_{t+4}$), respectively. The average warping error over the entire sequence of one video is calculated as:
\begin{equation}
	E_{warp}(V) = \frac{1}{ T-1 }\sum_{t=1}^{T-1} E_{warp}(V_{t},V_{t+G}).
\end{equation}

\begin{table}
	\centering
	\setlength{\tabcolsep}{2pt}
	\caption{\textbf{Short-range consistency.} We compare the short-range consistency using warping error($\downarrow$). {Bold entries denote the \textbf{Best} scores.}
	}\label{tab:short}
	\resizebox{\columnwidth}{!}{
		\begin{tabular}{l|ccccccc|c}
			\hline
			Method	&		BirdSeal		&		BoatSurf		&		MonkeySwan		&		DuckCrane		&		RobotCat		&		CraneSeal		&		RiderDeer		&		Average		\\
			\hline
			\multicolumn{9}{c}{Non-generative methods}\\
			\hline																																
			CutPaste	&		0.0025 		&		0.0055 		&		0.0028 		&		0.0009 		&		0.0003 		&		0.0004 		&		\textbf{0.0011} 		&		0.0019 		\\
			Poisson	&		\textbf{0.0014} 		&		0.0028 		&		0.0020 		&		0.0009 		&		0.0003 		&		0.0004 		&		0.0012 		&		\textbf{0.0013} 		\\
			Harmonizer	&		0.0025 		&		\textbf{0.0021} 		&		\textbf{0.0017} 		&		0.0009 		&		0.0003 		&		0.0004 		&		0.0013 		&		\textbf{0.0013} 		\\
			\hline
			\multicolumn{9}{c}{Generative methods}\\
			\hline																																
			TF-ICON	&		0.0034 		&		0.0064 		&		0.0036 		&		0.0020 		&		0.0023 		&		0.0131 		&		0.0045 		&		0.0051 		\\
			SVC	&		0.0007 		&		0.0016 		&		0.0022 		&		0.0017 		&		0.0008 		&		0.0008 		&		0.0029 		&		0.0015 		\\
			Ours	&	\textbf{	0.0002 	}	&	\textbf{	0.0010 	}	&	\textbf{	0.0022 	}	&	\textbf{	0.0014 	}	&	\textbf{	0.0003 	}	&	\textbf{	0.0003 	}	&	\textbf{	0.0016 	}	&	\textbf{	0.0010 	}	\\
			\hline
		\end{tabular}
	}
	\vspace{-2mm}
\end{table}

\begin{table}
	\centering
	\setlength{\tabcolsep}{2pt}
	\caption{\textbf{Long-range consistency.} We compare the short-range consistency using warping error($\downarrow$). {Bold entries denote the \textbf{Best} scores.}
	}\label{tab:long}
	\resizebox{\columnwidth}{!}{
		\begin{tabular}{l|ccccccc|c}
			\hline
			Method	&		BirdSeal		&		BoatSurf		&		MonkeySwan		&		DuckCrane		&		RobotCat		&		CraneSeal		&		RiderDeer		&		Average		\\
			\hline
			\multicolumn{9}{c}{Non-generative methods}\\
			\hline																														
			CutPaste	&		0.0028 		&		0.0046 		&		0.0036 		&		0.0012 		&		0.0032 		&		\textbf{0.0005} 		&		\textbf{0.0018} 		&		0.0025 	\\
			Poisson	&		\textbf{0.0018} 		&		0.0045 		&		0.0031 		&		0.0012 		&		0.0005 		&		0.0006 		&		0.0019 		&		0.0019 	\\
			Harmonizer	&		0.0028 		&		\textbf{0.0019} 		&		\textbf{0.0022} 		&		0.0012 		&		\textbf{0.0004} 		&		\textbf{0.0005} 		&		0.0021 		&		\textbf{0.0016} 	\\
			\hline
			\multicolumn{9}{c}{Generative methods}\\
			\hline																														
			TF-ICON	&		0.0040 		&		0.0084 		&		0.0044 		&		0.0025 		&		0.0028 		&		0.0016 		&		0.0059 		&		0.0042 	\\
			SVC	&		0.0009 		&		0.0029 		&	\textbf{	0.0029 	}	&		0.0020 		&		0.0010 		&		0.0010 		&		0.0036 		&		0.0020 	\\
			Ours	&	\textbf{	0.0004 	}	&	\textbf{	0.0010 	}	&		0.0033 		&	\textbf{	0.0018 	}	&	\textbf{	0.0006 	}	&	\textbf{	0.0005 	}	&	\textbf{	0.0023 	}	&	\textbf{	0.0014 	}	\\
			
			\hline
		\end{tabular}
	}
	\vspace{-2mm}
\end{table}

			

 The comparisons of short and long-range consistency are shown in Table.~\ref{tab:short} and  Table.~\ref{tab:long}, respectively. CutPaste, Poisson, and Harmonizer are non-generative methods, which essentially have better temporal consistency, but cannot produce interactive effects and are not harmonious; others and our method are generative methods, and the composited videos are more harmonious.  Nonetheless, the average of our metrics on temporal consistency is still superior all compared methods. 
\begin{figure}[t!]
	\centering
	\includegraphics[width=0.3\textwidth]{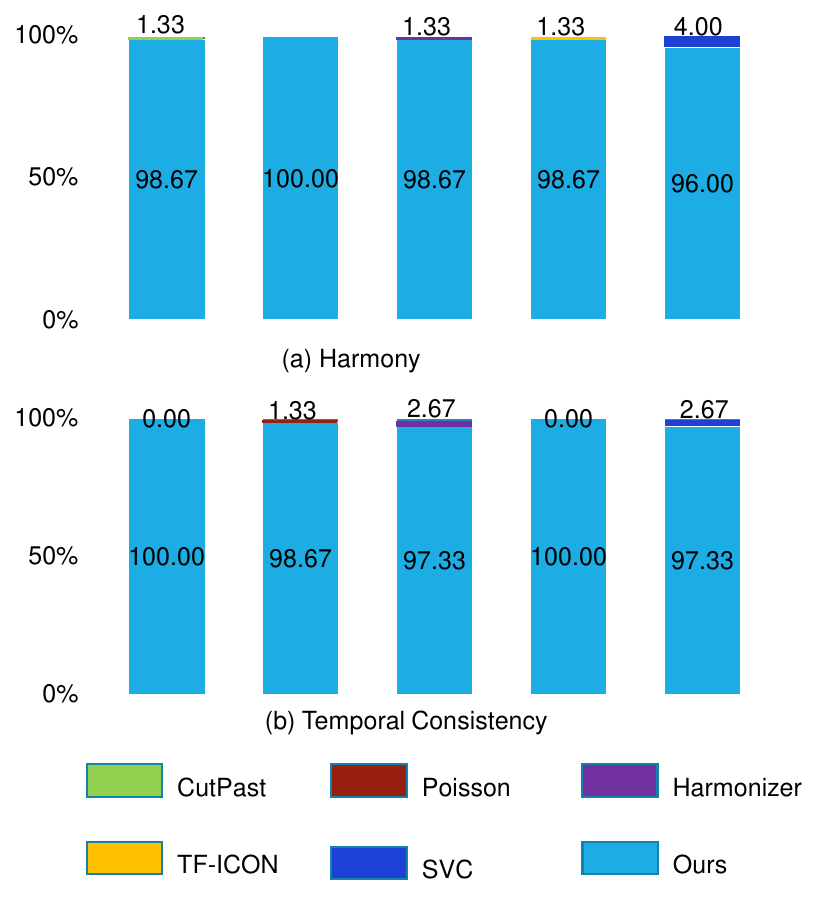}
	\caption{\textbf{User study.} Our results win more preferences both in the harmony and temporal consistency quality.}
	\label{fig:mvoc_userstudy}
\end{figure}

\begin{figure*}[t!]
	\centering
	\includegraphics[width=1.0\textwidth]{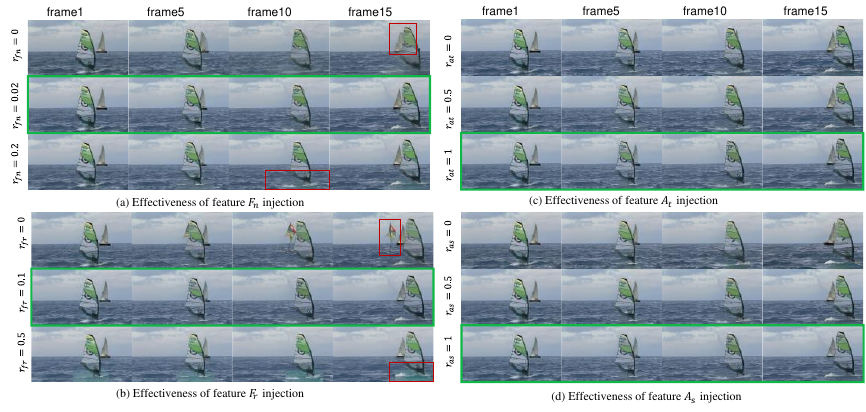}
	\caption{\textbf{Effectiveness of feature and attention injection.}}
	\label{fig:mvoc_ab_feature_attention}
	\vspace{-4mm}
\end{figure*}


\noindent\textbf{User study.} We show the advantage of our Multiple Video Object Composition method in real-world video composition through user studies. 25 participants are invited to vote on seven videos composited with different methods, i.e., CutPaste, Poisson, Harmonizer, TF-ICON, SVC, and ours.  Firstly, we showed the participants three input videos and two composited videos generated by our method and a compared method, and then recorded the number of participants voting on the harmony and temporal consistency of the composited videos. 
We collected 1750 votes for each of the evaluation indicators and presented the result in Fig.~\ref{fig:mvoc_userstudy} in the form of a box plot. It is clear that the proposed method performs well in terms of harmony and temporal consistency.

\subsection{Ablation study}
\label{sec:ablation_study_4.3}

 \begin{figure}[t!]
	\centering
	\includegraphics[width=0.45\textwidth]{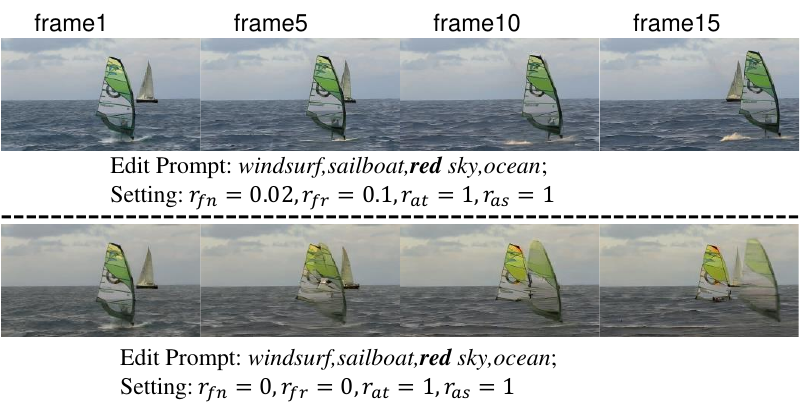}
	\caption{\textbf{Effectiveness of text guidance.}}
	\label{fig:mvoc_ab_test_guide}
	\vspace{-2mm}
\end{figure}

\begin{figure}[t!]
	\centering
	\includegraphics[width=0.45\textwidth]{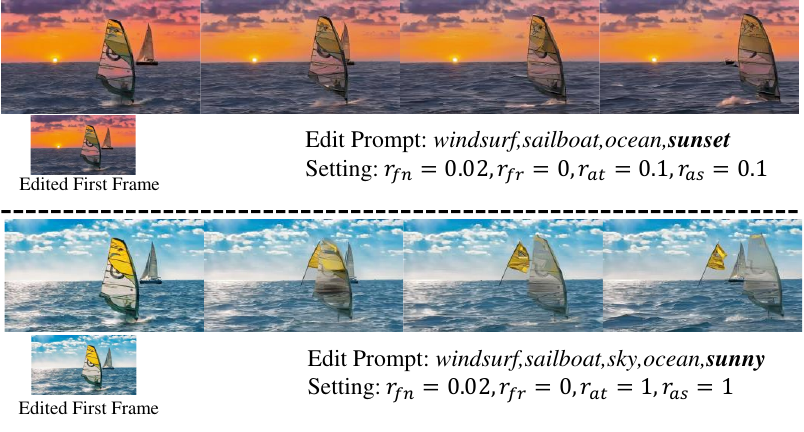}
	\caption{\textbf{Effectiveness of image guidance.}}
	\label{fig:mvoc_ab_generated_background}
	\vspace{-4mm}
\end{figure}

\noindent\textbf{Effectiveness of feature and attention injections.} Feature injection has a significant impact as evident from the potential distortions or generation of redundant objects in the composited video when the injection is absent. We have empirically set the hyper-parameters, i.e., $r_{fn}=0.02$, $r_{fr}=0.1$, $r_{at}=1$, $r_{as}=1$. We fixed three of the parameters while varying the remaining one to explore the parameter effects on the final composited video. Fig.~\ref{fig:mvoc_ab_feature_attention} shows the effectiveness of feature and attention injections. Feature injection demonstrates a notable influence, as it can lead to distortions or unnecessary object generation in the absence of such injection. Meanwhile, the feature injection intensity should not be too high, which leads to inharmonious in the composited video. The reason may be that the synthetic object is too biased towards the original video object. For example, in subfigure $(b)$, the water color below the sailboat is consistent with the original video. Attention injection has less effect on composited video than feature injection. However, attention injection will make the composited video more harmonious, including color consistency, etc.

\noindent\textbf{Effectiveness of text guidance.}
We use feature and attention injections to keep the objects in the composited video consistent with the motion and identity of the original objects, leading to the weak effect of text guidance. The effect of text guidance is only visible when the parameter injecting effect is reduced. Fig.~\ref{fig:mvoc_ab_test_guide} illustrates the results of this exploration. When using the default injection parameter , text guidance can not modify the sky color; while removing feature injection, text guidance produces some effects.


\noindent\textbf{Effectiveness of image guidance.} The compositing process is significantly influenced by the initial frame image, which allows for the modification of the objects or background depicted in the video. Fig.~\ref{fig:mvoc_ab_generated_background} illustrates the effect of modifying the composition of multiple video objects using the first frame image guidance. We use cosxl\footnote[1]{https://huggingface.co/spaces/multimodalart/cosxl} to modify the first image with the prompt of \emph{make sky to sunset} and \emph{make sky to sunny}, and then perform the generative process. It can be seen from the results that the composition effect is greatly affected by image guidance. However, when the hyper-parameter settings are not suitable, the model generates some unreasonable objects shown in the lower part of Fig.~\ref{fig:mvoc_ab_generated_background}.


\section{Limitations}
\label{sec:Limitations}
Our method has the following limitations: Firstly, our method needs to tune the hyper-parameters ( i.e.,  $r_{fn}$, $r_{fr}$, $r_{at}$, and $r_{as}$ ) when compositing video objects to produce better results, as discussed in \ref{sec:ablation_study_4.3}. Although the hyper-parameters empirically tuned can lead to good results in our experiments, it does not mean that these parameters can be applied to any videos. Secondly, the composited video objects need to keep the same camera motion as the original videos. In other words, if video objects have different camera motion directions, it may not be possible to composite them into a harmonious video. Finally, our method cannot modify the motion trajectory of a specified video object and control the viewpoint while compositing. These above limitations are exactly what video object composition methods need to address in future research.

\section{Conclusion}
\label{sec:Conclusion}
In this work, we propose a training-free multiple video object composition method that has the ability to composite video objects in a way maintaining the motion and identity consistency, while at the same time having the ability to generate interactive effects between different objects. We propose the video object dependence module that composites a harmonious video by injecting the features and attention maps of multiple video objects, which can be non-independent in the final video. In the future, we will expand our research to focus on reducing the number of hyper-parameters, compatibility of video object composition with different camera motions, and novel view editing of the video object.

%% file: sec/mvoc_derivation.tex
\section*{Supplementary Material}
\renewcommand\thesection{\Alph{section}}
\setcounter{section}{0}
\section{Derivation}
\label{sup:proof}
Given a set of objects $\{\boldsymbol{y}_1, \boldsymbol{y}_2, \ldots, \boldsymbol{y}_n\}$, the joint probability distribution of the composited objects $\{\boldsymbol{x}\}$ can be expressed as follows:
\begin{equation}
    \begin{split}
        p(\boldsymbol{x} |\boldsymbol{y}_1, \ldots, \boldsymbol{y}_n) &\propto p(\boldsymbol{x})p(\boldsymbol{y}_1, \ldots, \boldsymbol{y}_n|\boldsymbol{x})
    \end{split}
\end{equation}

The second term in above equation can be expressed as 
\begin{equation}
    \begin{split}
    p(\boldsymbol{y}_1, \ldots, \boldsymbol{y}_n|\boldsymbol{x})&=p(\boldsymbol{y}_1|\boldsymbol{x})p(\boldsymbol{y}_2, \ldots, \boldsymbol{y}_n|\boldsymbol{y}_1,\boldsymbol{x})\\
    &=\prod_{i=1}^np(\boldsymbol{y}_i|\boldsymbol{y}_0, \ldots, \boldsymbol{y}_{i-1},\boldsymbol{x})\\
    &=\prod_{i=1}^np(\boldsymbol{y}_i|\boldsymbol{x},\boldsymbol{y}_0, \ldots, \boldsymbol{y}_{i-1})\\
    &\propto \prod_{i=1}^n \frac{p(\boldsymbol{x} | \boldsymbol{y}_1, \ldots, \boldsymbol{y}_i)}{p(\boldsymbol{x} | \boldsymbol{y}_0, \ldots,\boldsymbol{y}_{i-1})},
    \end{split}
\end{equation}
where $\boldsymbol{y}_0 = \phi$, is the null condition.

Then we take a gradient of logarithm on both sides of Eq.\ref{eq:Socrefunction_bayes} w.r.t $\boldsymbol{x}$:
\begin{equation}
    \begin{split}
        \nabla_{x} \log p(\boldsymbol{x} |\boldsymbol{y}_1, \ldots, \boldsymbol{y}_n)&=\nabla_{x}\log p(\boldsymbol{x})\\
        &+\sum_{i=1}^n \bigl(\nabla_{x}\log p(\boldsymbol{x} | \boldsymbol{y}_1, \ldots, \boldsymbol{y}_i) \\
        &- \nabla_{x}\log p(\boldsymbol{x}|\boldsymbol{y}_0, \ldots, \boldsymbol{y}_{i-1}) \bigl) 
    \end{split}
\end{equation}

Finally, we may obtain a modified score prediction from the above expression as $\hat{\epsilon}_\theta(\boldsymbol{x}_t, t |\boldsymbol{y}_1, \ldots, \boldsymbol{y}_n)$, where $w_i$ controls the temperature of each implicit condition: 
\begin{equation}
	\begin{split}
    \hat{\epsilon}_{\theta}(\boldsymbol{x}_t, t |\boldsymbol{y}_1, \ldots, \boldsymbol{y}_n) &= \epsilon_\theta(\boldsymbol{x}_t, t) \\
    &+ \sum_{i=1}^n w_i\bigl(\epsilon_\theta(\boldsymbol{x}_t, t | \boldsymbol{y}_1, \ldots, \boldsymbol{y}_i) \\
    &- \epsilon_\theta(\boldsymbol{x}_t, t | \boldsymbol{y}_0, \ldots, \boldsymbol{y}_{i-1} )\bigl).
\end{split}
\end{equation}

\section{More comparisons}
\begin{figure*}[t!]
	\centering
	\includegraphics[width=0.9\textwidth]{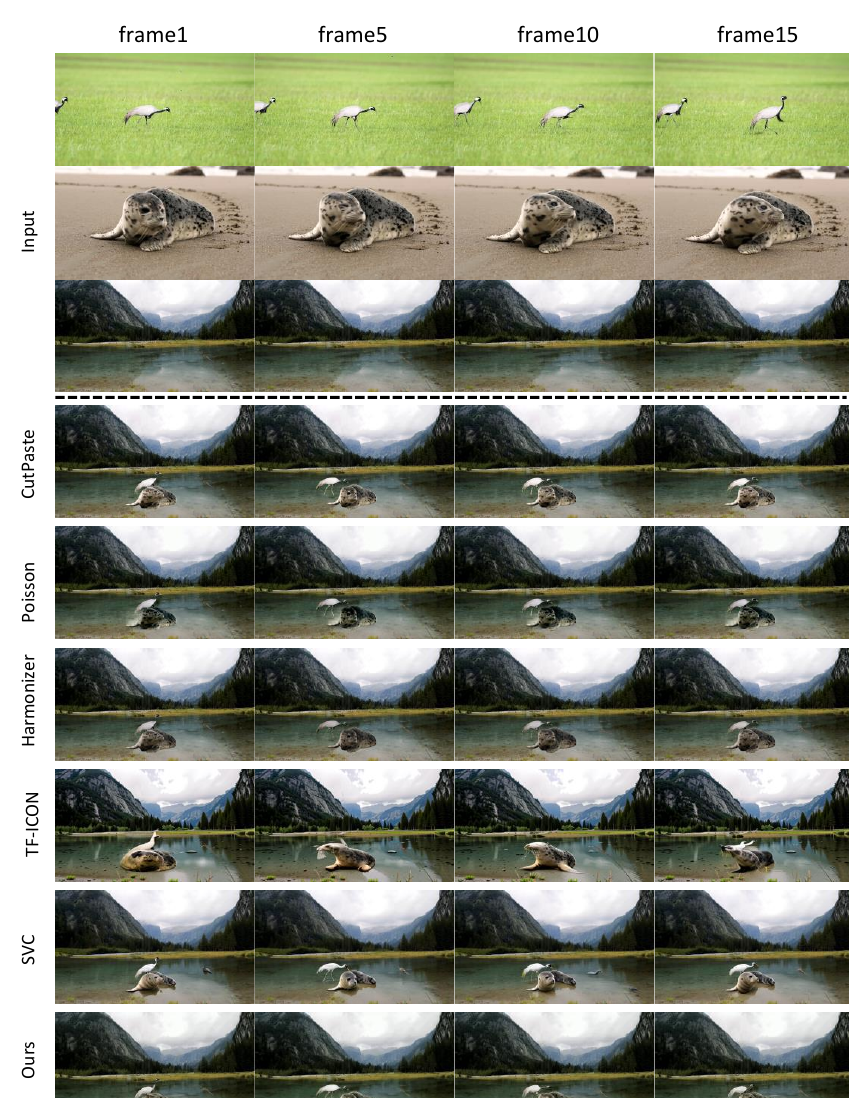}
	\caption{\textbf{Visual comparison on the video object of \emph{CraneSeal}.}}
	\label{fig:mvoc_supply_CraneSeal}
	\vspace{-4mm}
\end{figure*}

\begin{figure*}[t!]
	\centering
	\includegraphics[width=0.9\textwidth]{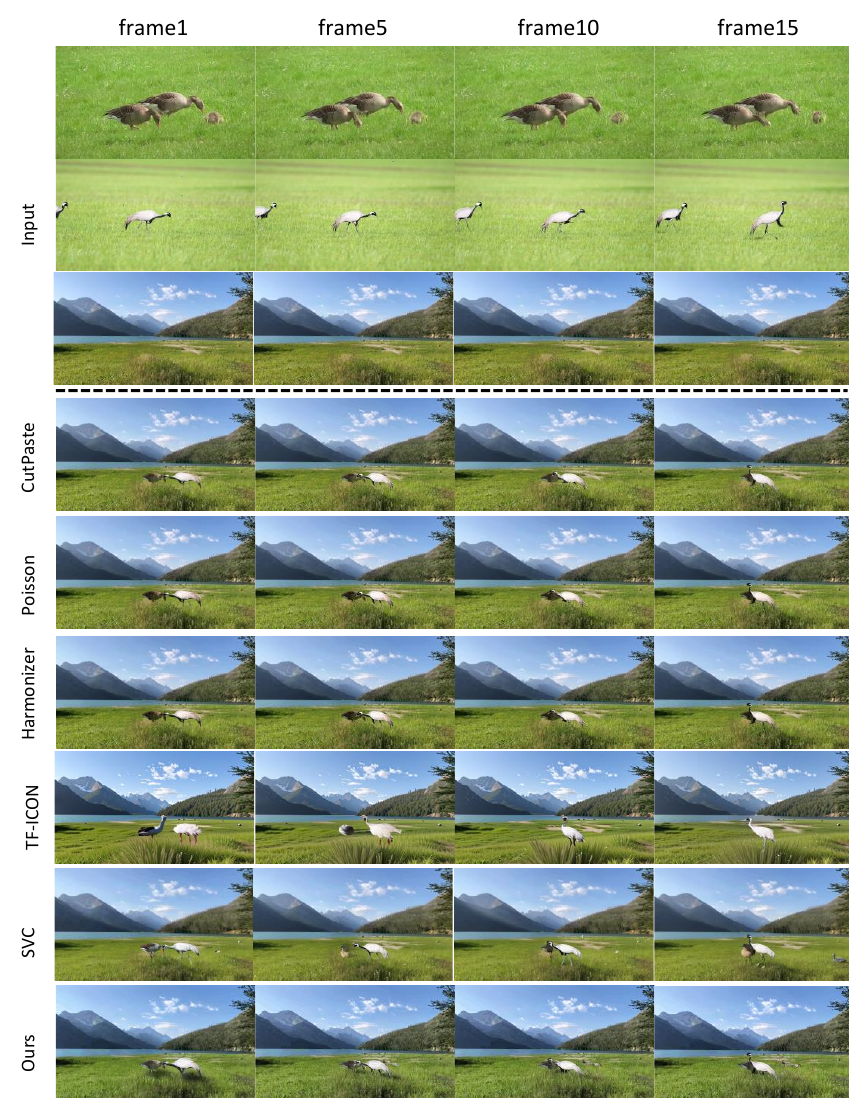}
	\caption{\textbf{Visual comparison on the video object of \emph{DuckCrane}.}}
	\label{fig:mvoc_supply_DuckCrane}
	\vspace{-4mm}
\end{figure*}

\begin{figure*}[t!]
	\centering
	\includegraphics[width=0.9\textwidth]{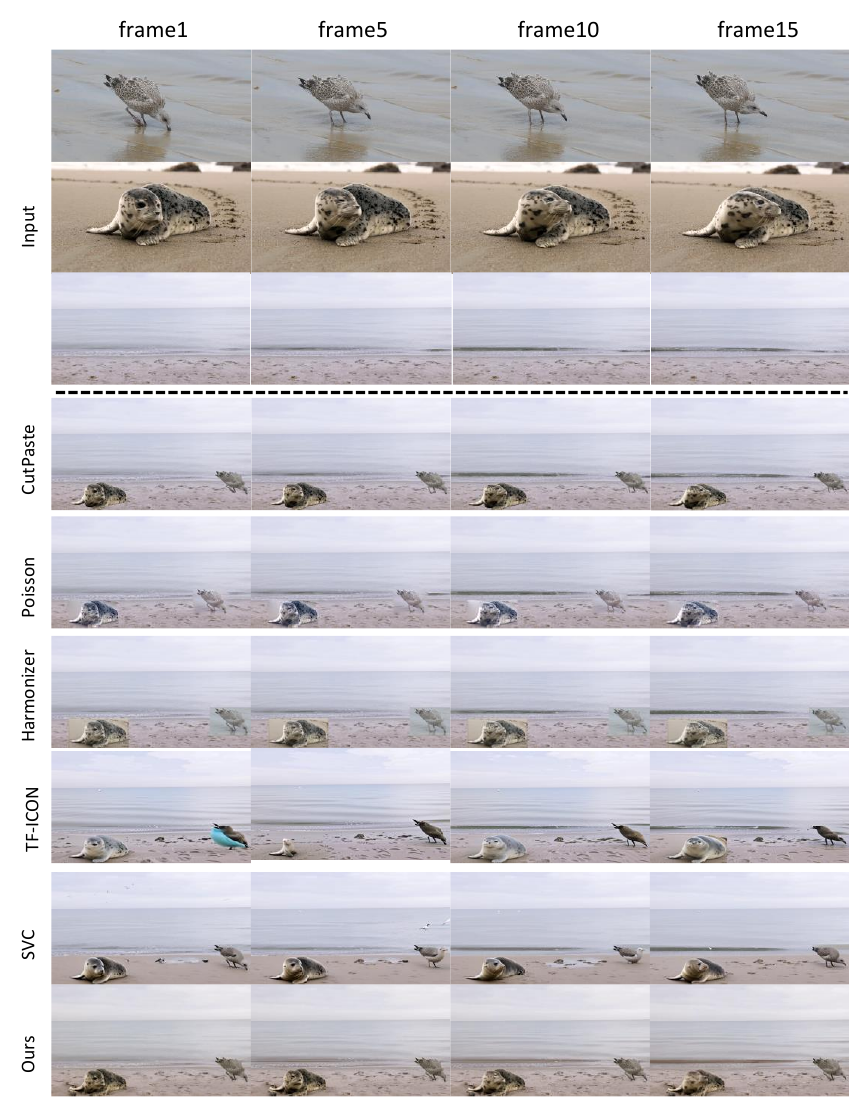}
	\caption{\textbf{Visual comparison on the video object of \emph{BirdSeal}.}}
	\label{fig:mvoc_supply_BirdSeal}
	\vspace{-4mm}
\end{figure*}

\begin{figure*}[t!]
	\centering
	\includegraphics[width=0.9\textwidth]{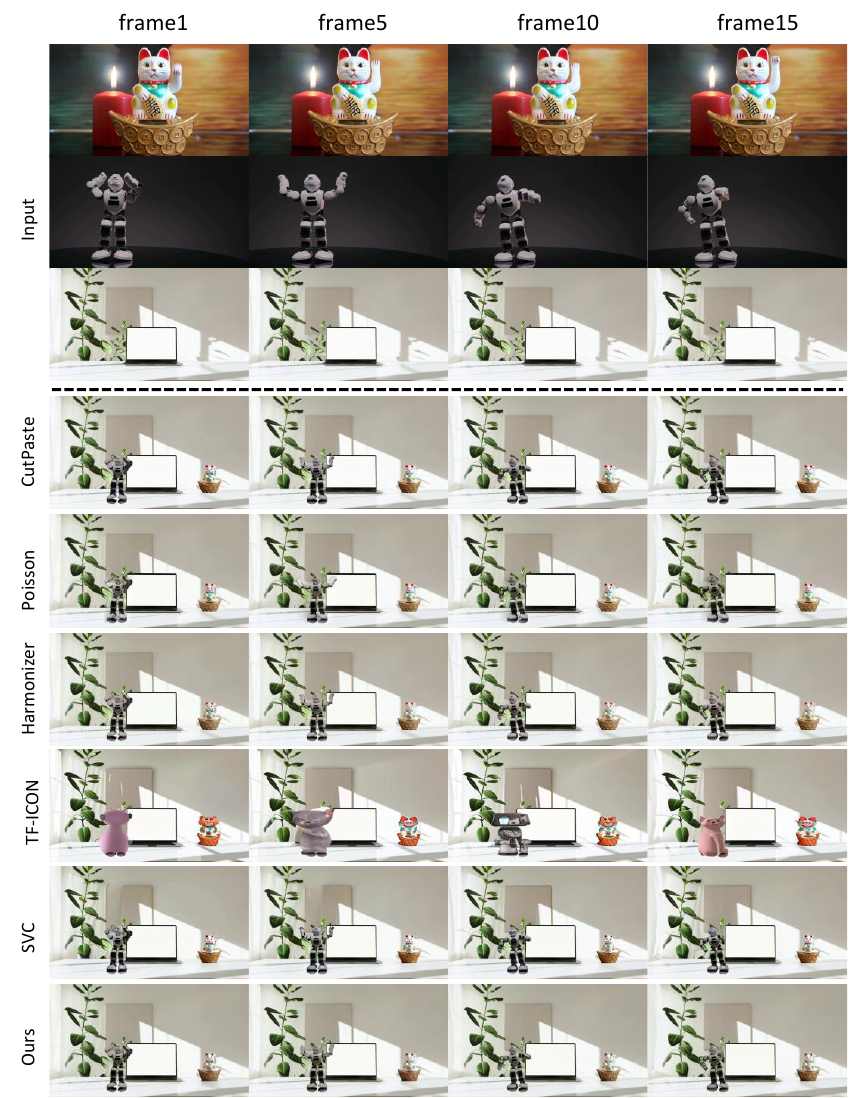}
	\caption{\textbf{Visual comparison on the video object of \emph{RobotCat}.}}
	\label{fig:mvoc_supply_RobotCat}
	\vspace{-4mm}
\end{figure*}

\begin{figure*}[t!]
	\centering
	\includegraphics[width=0.9\textwidth]{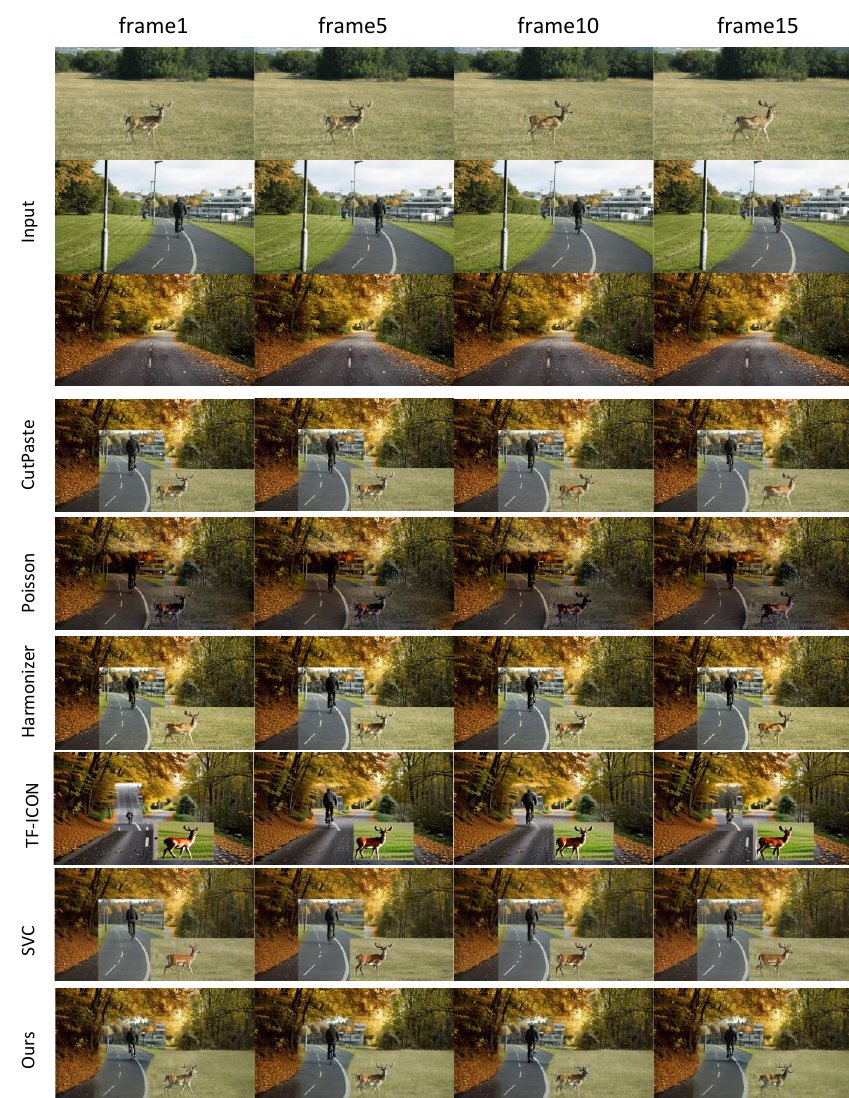}
	\caption{\textbf{Visual comparison on the video object of \emph{RiderDeer}.}}
	\label{fig:mvoc_supply_RiderDeer}
	\vspace{-4mm}
\end{figure*}

Please refer to Fig.~\ref{fig:mvoc_supply_CraneSeal},~\ref{fig:mvoc_supply_DuckCrane},~\ref{fig:mvoc_supply_BirdSeal},~\ref{fig:mvoc_supply_RobotCat} and Fig.~\ref{fig:mvoc_supply_RiderDeer} for more comparisons with the state-of-the-art methods.


